\documentclass[runningheads]{llncs}

\usepackage{times}
\usepackage{soul}
\usepackage{url}
\usepackage[hidelinks]{hyperref}
\usepackage[utf8]{inputenc}
\usepackage[small]{caption}
\usepackage{graphicx}
\usepackage{amsmath}
\usepackage{booktabs}
\usepackage[ruled, noline, linesnumbered]{algorithm2e}
\usepackage{subcaption}
\usepackage{tabularx}
\title{Explaining Deep Learning Models with Constrained Adversarial Examples}

\author{Jonathan Moore\inst{1} \and
	Nils Hammerla\inst{1} \and
	Chris Watkins\inst{2}}
\authorrunning{J. Moore et al.}
\institute{Babylon Health, London SW3 3DD, UK \email{\{jonathan.moore, nils.hammerla\}@babylonhealth.com} \and
	Royal Holloway University of London \email{c.j.watkins@rhul.ac.uk}}
\begin{document}

\maketitle

\begin{abstract}
	Machine learning algorithms generally suffer from a problem of explainability.
	Given a classification result from a model, it is typically hard to
	determine what caused the decision to be made, and to give an informative
	explanation. We explore a new method of generating counterfactual explanations, which instead of explaining why a particular classification was made explain how a different outcome can be achieved. This gives the recipients of the explanation a better way to understand the outcome, and provides an actionable suggestion. We show that the introduced method of Constrained Adversarial Examples (CADEX) can be used in real world applications, and yields explanations which incorporate business or domain constraints such as handling categorical attributes and range constraints.
	
\end{abstract}

\section{Introduction}
The recent explosion in the popularity of machine learning methods has led to their wide adoption in various domains, outside the technology sector. Machine learning algorithms are used to predict how likely convicted felons are to recidivate, which candidates should be interviewed for a job, and which bank customers are likely to default on a given loan. These algorithms assist human decision making, and in some cases may even replace it altogether. When humans are responsible for a decision, we can ask them to explain their thought process and give a reason for the decision (although often that is not done). Asking a machine learning algorithm to explain itself is a challenging problem, especially in the case of deep neural networks.

Throughout this work, we will refer to the following scenario. Assume that a bank has trained a deep learning model to predict which of its customers should be eligible for a loan. The input is a vector that represents the customer, using attributes such as age, employment history, credit score, etc. The output is a label which says whether said customer is likely to repay a loan or default. Now suppose that a customer requests a loan, and is denied based on the decision of the algorithm. The customer would obviously like to know why he or she was rejected, and what prompted the decision. The bank, on the other hand, is faced with two problems:
\begin{itemize}
	\item The bank has difficulty giving a meaningful explanation. Various explanation methods exist, but it is hard to determine which ones give valuable feedback to the customer.
	\item The bank doesn't want to expose its algorithm, or even the full set of features it uses for classification. Credit scoring and loan qualification mechanisms are typically closely guarded by most banks.
\end{itemize}

The Constrained Adversarial Examples (CADEX) method presented here aims to answer both problems. Instead of directly explaining why a model classified the input to a particular class, it finds an alternate version of the input which receives a different classification. In the bank scenario it produces an alternate version of the customer, which would get the loan. The customer can act on this explanation in order to receive a loan in the future, without the bank revealing the inner working of its algorithms. Such explanations are referred to as \textit{Counterfactual 
Explanations}. As shown in a recent study of AI explainability from the perspective of social sciences by \cite{Miller:2019}, people tend to prefer contrastive explanations over detailed facts leading to an outcome, and that they find them more understandable. In fact, when people explain why an event occurred, they tend to explain it in comparison to another event which did not occur.

The CADEX method offers several improvements over current techniques for finding counterfactual explanations:

\begin{itemize}
	\item It supports directly limiting the number of changed attributes to a predetermined amount.
	
	\item It allows specifying constraints on the search process, such as the direction attributes are allowed to change. 
	
	\item It fully supports categorical one-hot encoded attributes and ordinal attributes.
	
	\item It surpasses current explainability methods by providing better, more understandable explanations.
\end{itemize}
\section{Related Work}
\subsection{Explainability} \label{sec:explainability}
As machine learning models become increasingly complex and have a large number of internal weights and dependencies, it becomes more and more challenging to explain how they work, and why they produce the predictions they make. Explainability of machine learning models has been an active topic of research recently, and multiple methods and techniques have been developed to try and address these difficulties from several points of view.

Some methods attempt to explain what a model has learned in its training phase. Such methods examine the weights of the trained model and present them in an interpretable way. These methods are particularly common for CNNs, and so the explanations have a highly visual nature. A recent survey \cite{Zhang:2018} mentions many such techniques, which include visualizing the patterns learned in each layer and generating images which correspond to feature maps learned by the network. However, these methods are all specifically tailored to work on CNNs and don't generalize to any black box model.

Other methods seek to explain the output of a classifier for a specific given input. These methods aim to answer the question: ``why did the model predict this class?", by assigning a weight or significance score to the individual features of the input. Most notable in this category are LIME \cite{Ribeiro:2016} and SHAP \cite{Lundberg:2017}.

LIME takes a given input, and creates different versions of it by zeroing various attributes (or super pixels in the case of images), and then builds a local linear model while weighting the inputs by their distance to the original. The model is trained to minimize the number of non-zero coefficients by using a method such as LASSO. This results in an explainable linear model where the model's coefficients act as the explanation, and describe the contribution of each attribute (or super pixel) to the resulting classification.

SHAP attempts to unify several explanation methods such as LIME and DeepLift \cite{Shrikumar:2017}, in a way that the feature contributions are given in Shapley values from game theory, which have a better theoretic grounding than those produced by LIME. For tabular data the method is called Kernel SHAP, which improves LIME by replacing the heuristically chosen loss function and weighting kernel with ones that yield Shapley values.

Both methods produce an output that highlights which attributes contributed most to the classification, and which reduced the probability of classification. There are several drawbacks to this approach. First, it typically requires a domain expert to understand the significance of the output, and what the values mean for the model. Second, the explanation they provide is not actionable. In the bank scenario, they can tell the user, for example, that she didn't get the loan because of her salary and age. They won't say what she needs to do to get the loan in the future - should she wait until she's older? How much older? Or can she change another attribute such as education level and get the loan?

Some recent methods try to provide such explanations by looking for \textit{counterfactuals}. A counterfactual explanation answers the question: ``Why was the outcome $Y$ observed instead of $Y^*$?". The more specific formulation for machine learning models is: ``If $X$ had the values of $X^*$, the outcome $Y^*$ would have been observed instead of $Y$", where $X$ represents the input to the model. By observing the difference between $X$ and $X^*$, we can provide a ``what-if" scenario which is actionable to the end user. In the bank scenario the explanation could be, for example, ``If you had \$5000 instead of \$4000 in your account you would have gotten the loan".

A na\"ive way of finding counterfactuals would be to simply find the nearest training set instance to the input, which receives a different classification. The limitation in that approach is that it is limited by the size and quality of the training set. It cannot find a counterfactual that isn't explicitly in the set. Additionally, showing the user a counterfactual which represents the details of another user may not even be legal considering data protection rights and confidentiality.

\cite{Laugel:2018} finds synthetic counterfactual explanations by sampling from a sphere around the input in a growing radius, until one is found which classifies differently than the original. Then, the number of changed attributes is constrained by iteratively setting them to the value of the original as long as the same contrastive classification is kept. However, the method is sensitive to hyperparameter choices which affect how close the found counterfactuals will be to the original, and the paper doesn't suggest how to determine their optimal value.

\cite{Wachter:2018} generates counterfactuals by optimizing a loss function, which factors the distance to the desired classification as well as a distance measure to the original input. The distance measure is used to limit the number of attributes changed via regularization, but the process of finding the counterfactual requires iteration over various coefficient values, and doesn't allow a hard limit on the number of changed attributes. In addition, it doesn't have a facility to handle one-hot encoded categorical attributes.

\subsection{Adversarial Examples}
Adversarial examples were discovered by \cite{Szegedy:2014}, who showed that given a trained image classifier, one could take a correctly classified image and perturb its pixels by a small amount which is indistinguishable to the human eye, and yet causes the image to receive a completely different label by the classifier. \cite{Goodfellow:2015} developed an efficient way of finding adversarial examples called FGSM, which uses the neural network's loss function's gradient to find the direction where adversarial examples can be found. \cite{Kurakin:2016} improved the technique and enabled it to target a specific desired classification, as well as using a more iterative approach to find the adversarial examples.

Most of the discussion around adversarial examples has been in the context of security and attacks against models deployed for real world applications.  \cite{Papernot:2017} demonstrate an attack against an online black-box classifier, by training a different classifier on a synthetic dataset and showing that adversarial examples found on that classifier also fooled the online one. They also show that multiple types of classifiers can be attacked that way, such as linear regression, decision trees, SVMs, and nearest neighbours. Others show that adversarial examples can carry over to the real world by printing or 3D printing them, and fooling camera based classifiers (\cite{Kurakin:2016,Brown:2017,Athalye:2017}).

CADEX uses adversarial examples to facilitate an understanding of the model instead of attacking or compromising it, by finding counterfactual explanations close to the original input. The search process is constrained to enforce domain or business constraints on the desired explanation.

\section{Generating Explanations}
We present the CADEX method for generating explanations for deep learning models. Let $f(x)=\hat{y}$, where $f$ is the model, $x$ is a specific input sample, and $\hat{y}$ is the output class. The method aims to find $x^*$ for which $f(x^*)=y^*$, where $\hat{y} \ne y^*$, and $x^*$ is as close as possible to $x$ while satisfying a number of constraints. This allows us to present the user a ``what if" scenario. In the case of the bank loan application, the user can be told: ``if you had the attributes of $x^*$, you would get the loan". That is, in that scenario $\hat{y}=\textsc{reject}$ and $y^*=\textsc{approve}$. The full algorithm is listed as algorithm \ref{alg:cadex}. The code used to implement the method and perform the evaluation can be found at \texttt{https://github.com/spore1/cadex}

\subsection{Finding Adversarial Examples}
The main motivation in CADEX is to find the explanations through adversarial examples. Adversarial examples work in a very similar sense, by changing the model input with a minimal perturbation so that it receives a different classification. 

%The main motivation in CADEX is to find the explanations through adversarial examples. Adversarial examples work in a very similar sense, by changing the model input so that it receives a different classification. Multiple methods have been proposed for this purpose, the most well known being the Fast Gradient Sign method (FGSM) \cite{TODO}, which simply adds the gradient of the model's loss function to the input, multiplied by a constant factor. While this method is efficient and effective, it does suffer from a number of drawbacks. First, it doesn't target a particular desired output classification, it simply attempts to find a different classification than the original. Second, it doesn't guarantee that an adversarial example is actually found, since the chosen factor may be too small. Similarly, the factor may be too large, and we wouldn't get an adversarial example with the smallest total change in attributes. Some extensions to the FGSM method have been proposed (for example, \cite{TODO}), which turn it into an iterative method and also allow specifying a target classification. The method used in our implementation of CADEX takes a slightly different approach, which generates adversarial examples that are closer to the original input.

Given the original input $x$, we can calculate the loss of the model between the actual output $\hat{y}$, and the desired target classification $y^*$. This is typically the cross entropy loss between the predicted class probabilities and the desired one. Then, we take the gradient of the loss with respect to the input.% 
\begin{align}\label{eq:gradient}
    \nabla Loss = \frac{\partial}{\partial{x}}Loss(\hat{y}, y^*)
\end{align}%
We then follow the gradient in input space using an optimizer such as Adam (\cite{Kingma:2014}) or RMSProp, until $f(x^*)=y^*$, which is our target classification. This typically results in an input that lies right on the decision boundary between the classes.

\subsection{Constraining the Number of Changed Attributes}\label{sec:changed_attr}
Following the method above will indeed find adversarial examples that are close to the original sample. However, since we calculate the gradient in input space and don't constrain it, any number of attributes in $x$ may change. Typically, the gradient is nonzero for all attributes, meaning that the resulting $x^*$ is different than $x$ in all attributes. The issue with this approach is that there could be dozens or hundreds of different attributes, and showing the user an explanation which is different in so many attributes is hardly useful, and doesn't constitute an explanation the user can act on. Ideally, we would like to limit the changed attributes to a small number, so that it is perceived as actionable by a human.

Previous work such as \cite{Wachter:2018} has attempted to limit the number of changed attributes by adding a form of L1 regularization to the loss function. However, this approach cannot guarantee the number of changed attributes will in fact be under an acceptable amount.

We limit the number of changed attributes by applying a mask to the gradient. The mask is used to zero the gradient in all attributes except the ones we wish to allow to change. When this gradient is applied to the input, only the selected attributes will be modified. The decision of where to zero out the gradient is performed as follows. First, get the gradient of the loss function with respect to the input as in equation \ref{eq:gradient}. Then, sort the gradient attributes by their absolute value from large to small, and take the top $n_{change}$ attributes, where $n_{change}$ is the number of desired attributes to change (e.g. 3 or 5). Then, prepare a mask which is set to 1 for the top $n_{change}$ attributes and zero elsewhere. At each iteration of gradient descent, after getting the gradient but before applying it to the weights by the optimizer, multiply the gradient by the mask. Then proceed as usual to update the weights.

\subsection{Constraining the Direction of the Gradient}
In addition to the number of changed attributes, we may want to place another constraint on the search process. The gradient may change each attribute in any direction - positive or negative - which depending on context may not be acceptable. Consider that in the bank scenario, the input may contain attributes such as ``age" or ``number of children". The algorithm may suggest that the user would get the loan if she were younger, or if she had one less child. For obvious reasons, no bank would ever want to make such a suggestion. We therefore wish to constrain the direction that some of the attributes would be permitted to change in. 

We introduce a new parameter $C$ which is used to build a mask to further constrain the gradient. This parameter is a vector of the same dimensions as the model's input, and is defined to be positive for each attribute that may only increase in value, negative for those that may only decrease, and 0 where the value may go in any direction. We assume that this will be defined by a domain expert, who understands each attribute in the data and the implications of changing it. Then, we build the following mask, for each attribute $i$ in the input vector:%
\begin{align}
 C_{{mask}_i} = & \begin{cases}
 	1 & \text{if } C_i > 0 \text{ and } \nabla Loss_i < 0 \text{ or}\\
 	 & C_i < 0 \text{ and } \nabla Loss_i > 0 \\ 	
 	0 & \textrm{else}
 \end{cases}
\label{eq:direction_mask}
\end{align}%
Note that if $C$ is \textit{positive}, we allow only a \textit{negative} gradient, since the gradient is subtracted from the current input at each step of gradient descent, and vice versa when $C$ is negative.

The resulting mask is used during training similar to the process described in section \ref{sec:changed_attr}. In fact, the two techniques can be used together, by first using the directional constraint function to zero the gradient where needed, followed by selecting the top $n_{change}$ attributes. This way, the selected attributes are those which change in the allowed direction.

\subsection{Handling Categorical and Ordinal Attributes}
\begin{figure}[t]
	\centering
	\begin{subfigure}[b]{0.35\textwidth}
		\centering
		\includegraphics[scale=0.55]{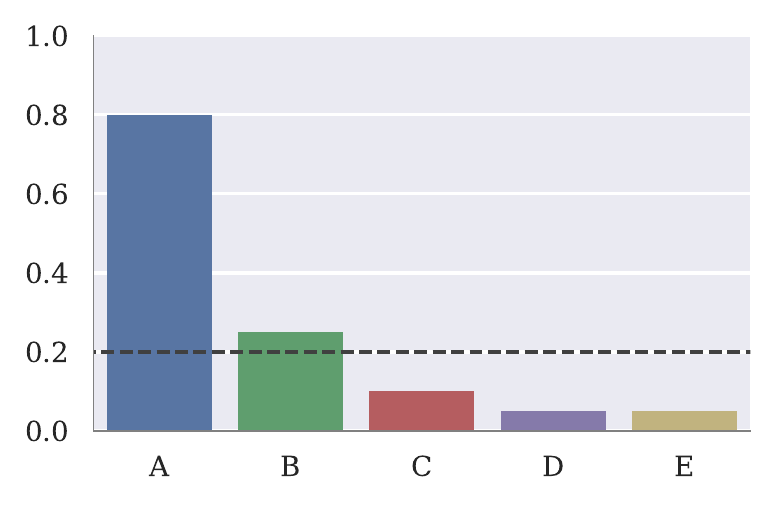}
		\caption{Before adjustment}
	\end{subfigure}%
	\begin{subfigure}[b]{0.35\textwidth}
		\centering
		\includegraphics[scale=0.55]{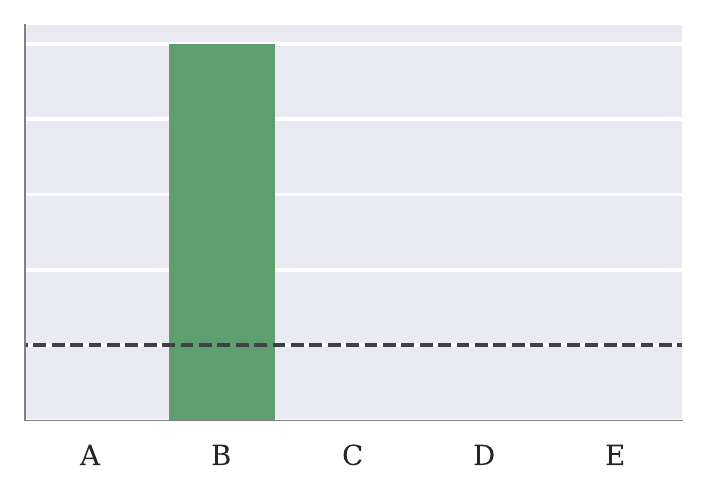}
		\caption{After adjustment}
	\end{subfigure}%
	\caption{Visual explanation of categorical attribute adjustment. On the left, the internal state of the algorithm has category A with the highest value, followed by B. We assume that A is the categorical value of the original sample. Category B is above the predefined threshold of 0.2, so its attribute is set to one and the rest are zeroed.}
	\label{fig:adjustcat}
\end{figure}
\begin{algorithm}[t]
	\caption{\textsc{Cadex} - counterfactual explanation for a given input}
	\label{alg:cadex}
	\DontPrintSemicolon
	\KwIn{
		$x$: original input sample\newline
		$f(x)$: trained model \newline
		$target$: desired output class for input $x$ \newline
		$max\_epochs$: maximum number of epochs to allow \newline
		$n_{change}$: maximum number of changed attributes \newline
		$C$: directional constraints \newline
		$n_{skip}$: number of attributes to skip from the top \newline
		$t_{flip}$: threshold to flip categorical attributes
	}	
	\KwOut{$x^*$: modified input sample $x$ which classifies as $target$}
	%	\tcc{Initialization:}
	$x^* \gets x$\; 
	$\nabla_{0} \gets \frac{\partial}{\partial x} Loss(f(x), target) $\; 
	$\nabla_{0} \gets \nabla_{0} * C_{mask}  $  \tcp{See eq \ref{eq:direction_mask}} 
	$i \gets$ \textsc{Argsort}($\nabla_{0}$) in descending order \; 
	$mask \gets 0$ \; 
	$mask\big[i[n_{skip}..n_{skip}+n_{change}]\big] \gets 1$ \; 
	%	\tcc{Training:}
	$result \gets \emptyset;\ epoch \gets 0$ \;
	\While{$epoch<max\_epochs$ \textup{\textbf{and}} $result=\emptyset$}{ 
		$\nabla_{epoch} \gets \frac{\partial}{\partial x} Loss(f(x), target)*mask$\; 
		$x^* \gets x^*-$\textsc{Adam(}$\nabla_{epoch}$)\; 
		$x^* \gets$ \textsc{FlipCategorical($x^*, t_{flip}$)}\; 
		$x_{adjusted} \gets$ \textsc{ApplyConstraints($x^*$)} \; 
		\If{f($x_{adjusted}$)=$target$}{ 
			$result \gets x_{adjusted}$\;
		}
		$epoch \gets epoch+1$\;
	}
	\Return{$result$}
	
\end{algorithm}
\begin{algorithm}[t]
	\caption{\textsc{FlipCategorical}}
	\label{alg:flipcat}
	\DontPrintSemicolon 
	\KwIn{$x^*$: modified input sample\newline
		$t_{flip}$: threshold
	}
	\KwOut{$x^*$ with flipped attributes where neccessary}
	$result \gets x^*$\;
	\ForEach{attr\_set \textup{\textbf{in} categorical attributes of} $x^*$} {
		$i \gets$ \textsc{Argsort}($x^*[attr\_set]$) in decreasing order\;
		\If{$x^*[i[1]]>t_{flip}$}{
			$result[attr\_set] \gets 0$\;
			$result[i[1]] \gets 1$\;
		}
	}
	\Return{$result$}\;
\end{algorithm}

\begin{algorithm}[t]
	\caption{\textsc{ApplyConstraints}}
	\label{alg:clipcat}
	\DontPrintSemicolon 
	\KwIn{$x^*$: modified input sample}
	\KwOut{$x^*$ with adjusted attributes}
	$result \gets x^*$\;
	\ForEach{attr\_set \textup{\textbf{in} categorical attributes of} $x^*$} {
		$i \gets$ \textsc{Argmax}($x^*[attr\_set]$)\;
		$result[attr\_set] \gets 0$\;
		$result[i]\gets 1$\;
	}
	\ForEach{attr \textup{\textbf{in} ordinal attributes of $x^*$}}{
		$result[attr]\gets \textsc{Round}(result[attr])$ \;
	} 
	\Return{$result$}\;
\end{algorithm}

Categorical attributes are frequently found in many real world datasets. They pose a challenge to the algorithm, which relies on changing the attributes gradually by following the gradient. Categorical attributes are typically one-hot encoded, which means that each attribute may only be set to 0 or 1, and only one attribute per attribute set must be set to 1 at any given time. By na\"ively following the gradient, the algorithm will easily violate these constraints.

We use the following method to deal with categorical attributes. During training, we continue to treat the categorical attributes as any other attribute, in the sense that they are allowed to change gradually by the gradient. Internally, there could be a moment where the representation of the modified input sample violates the rules of one-hot encoded attributes, but that is acceptable as long as we don't return this as the final result. At each epoch, two extra steps are performed. First, a check is made to determine whether certain categorical attributes need to be ``flipped", that is to set the value 1 to a different category than that of the original. Assuming that an ``attribute set" is defined to be the set of attributes that represent a one-hot encoded categorical value, then for each attribute set we find the second highest valued attribute, and if it's above a threshold $t_{flip}$ we set it to 1 and zero the rest. The reasoning behind this is that the highest attribute would be that which was equal to one in the original sample, and the second highest is the one that has been most affected by the gradient. This is illustrated visually in figure \ref{fig:adjustcat} and described in algorithm \ref{alg:flipcat}. The threshold is a hyper-parameter which tunes how quickly the algorithm choses to change the categorical attributes.

Additionally, at each epoch we need to determine whether the stopping condition has been met, which is that the modified observation is classified as the desired label. We test this against an adjusted version of the observation, where the highest attribute in each attribute set to one and the rest to zero. This means we're testing against the valid version of the observation, where attribute values can only have values of 1 or 0. 

Ordinal attributes - which must hold integer values - are handled in a similar fashion. During training they are allowed to have any fractional value, but when evaluating the stopping condition we round them to the nearest integer. The adjustment process is described in algorithm \ref{alg:clipcat}.

\subsection{Finding Alternate Explanations}
In some cases, it would be useful to be able to present more than one adversarial example and show user multiple alternate scenarios with the desired classification. 

As described in section \ref{sec:changed_attr}, the method sorts the gradient in descending order of the absolute value of the attributes, and selects the top $n_{change}$ attributes. By skipping the first top $n_{skip}$ attributes, the method chooses a different set of attributes to change and will arrive at a different solution. Thus, by trying various values for $n_{skip}$, we can generate multiple alternate adversarial examples.

\section{Evaluation}

%\begin{figure}[t]
%	\centering
%	\begin{minipage}{0.49\textwidth}
%		\includegraphics[width=\linewidth]{figures/number_solutions.pdf}
%		\caption{Number of solutions found by $n_{changed}$}
%		\label{fig:number_solutions}
%	\end{minipage}
%	\begin{minipage}{0.49\textwidth}
%		\includegraphics[width=\linewidth]{figures/distance_to_training.pdf}	
%		\caption{Cummulative distribution of distances found using CADEX vs. training set}
%		\label{fig:distance}
%	\end{minipage}
%\end{figure}
%

We evaluate CADEX with several different approaches. First, we train a feed-forward neural network on the German loan dataset \cite{Dua:2017}, which contains 1000 observations of people who applied for a loan and has a range of numeric, categorical and ordinal attributes. Every categorical attribute was one-hot encoded, and assigned a readable label from the data dictionary supplied with the data. Numerical attributes were normalized to have zero mean and unit standard deviation. The dataset was split to 80\% training and 20\% validation. The model had one hidden layer with 15 neurons with ReLU activations, and one classification layer with two output neurons with softmax activations to represent the classification labels of \textsc{Approved} and \textsc{Rejected}. The model was trained using the Adam optimizer with early stopping when the validation loss started to increase.

Then for each training set sample, we ran CADEX to find 10 explanations by varying $n_{skip}$ from 0 to 9. We used ordinal constraints on the attributes \textsc{Existing\_credits} and \textsc{People\_maintained}, and directional constraints to allow only positive changed on \textsc{Age} and \textsc{People\_maintained}. $t_{flip}$ was set to 0.2 for all experiments. We repeated the above process for $n_{changed}=(5, 7, 10)$.

\subsection{Sample Explanation}
Table \ref{tbl:samples} Illustrates three explanations found for one particular validation set sample, who was refused a loan. The explanations are clear and concise, and can be immediately understood by non-domain experts. They also provide an interesting insight into the inner workings of the model and what it has managed to learn. We can see that the individual would have been given a loan if she were older or had a longer employment history. We can also see it would have been better for her not to have a checking account at all rather than have a negative balance. Finally, we learn that had she been a male instead of a female, she would have gotten the loan which indicates a possible bias of the model to prefer men over women.

Upon investigation, we found that for all of the women which the model classified as \textsc{Reject}, we were able to produce a counterfactual that changed the sex attribute to male, and therefore the model is in fact biased. We conclude that in addition to providing actionable explanations to an end user, CADEX is also a valuable method to aid in the understanding the inner workings of the model.

\begin{table*}[t]
	\begin{tabularx}{\linewidth}{Xp{7.5em}lll}
		\toprule
			\textbf{Attribute} & \textbf{Original} & \textbf{Explanation 1} & \textbf{Explanation 2} & \textbf{Explanation 3} \\
		\midrule
			duration & 24 & 21.74 & - & - \\
			credit & 3123 & 2563.85 & - & - \\
			installment percent & 4 & 3.77 & 3.59 & - \\
			age & 27 & 29.31 & 31.09 & - \\
			account status & $<$ 0 DM & no checking account & - & - \\
			sex status & female  & - & male single & - \\
			property & building society \newline savings agreement & - & real estate & - \\
			employment & $<$ 1 year & - & 4..7 years & 4..7 years \\
			purpose & car (new) & - & - & car (used) \\		
		\bottomrule
	\end{tabularx}
	\caption{Sample explanations for one refused loan candidate}
\end{table*}
\label{tbl:samples}

\begin{figure}[t]
	\centering
	\includegraphics[scale=0.8]{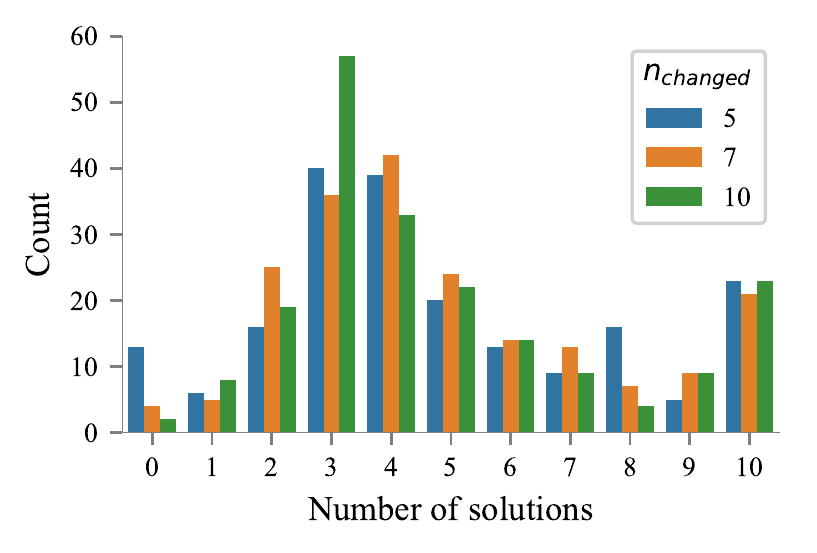}
	\caption{Number of solutions found by $n_{changed}$}
	\label{fig:number_solutions}
\end{figure}	
\begin{figure}[t]	
	\centering
	\includegraphics[scale=0.8]{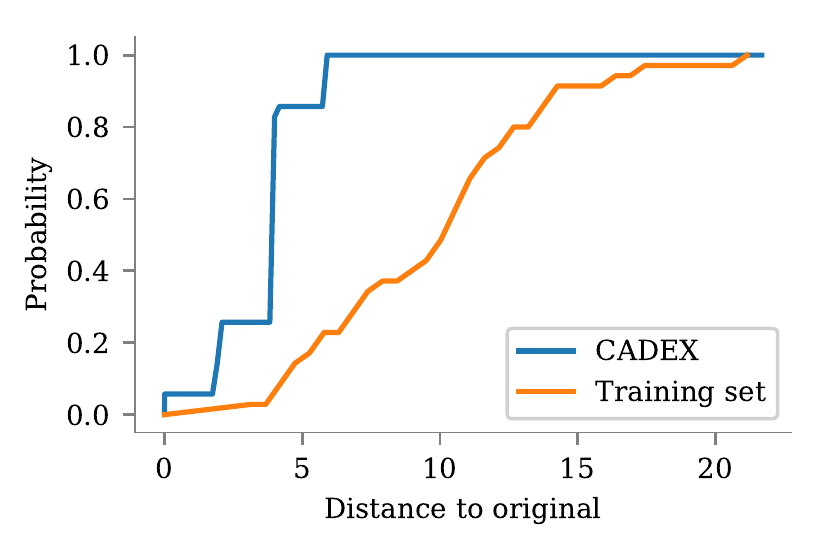}	
	\caption{Cummulative distribution of distances found using CADEX vs. training set}
	\label{fig:distance}
\end{figure}
\begin{figure}[t]
	\centering
	\hspace{-1cm}
	\includegraphics[scale=0.8]{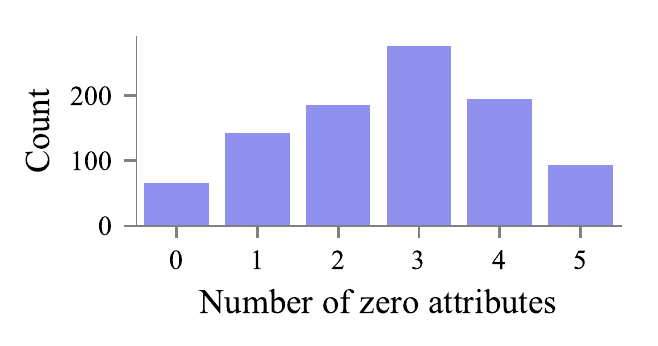}
	\caption{Distribution of zero SHAP attributes, which were used to produce counterfactual explanations by CADEX}
	\label{fig:shap}
\end{figure}
\subsection{Number of Solutions Found}
It is possible that for a particular configuration of CADEX parameters, the method will not converge on an adversarial example. Since we're zeroing many of the gradient's elements, it may get stuck in a local minimum or simply not point at the right direction to cross the decision boundary. To see how significant this is, we plot histograms of how many solutions were found per training set item, for the 3 values of ${n_{changed}}$. As can be seen in figure \ref{fig:number_solutions}, for most samples CADEX finds at least 3 or 4 explanations which should be enough for any real world use case.

\subsection{Comparison to Training Set Counterfactuals}
We compared CADEX to the method of finding the counterfactuals directly from the training set. For each item in the validation set which was denied a loan, we find nearest training set sample using L2 distance which receives a different classification, without limiting the number of attributes that are allowed to change. We plot the cumulative distribution of the distances compared with those found using CADEX. As can be seen from figure \ref{fig:distance}, CADEX generates counterfactual explanations that are much closer to the original.

\subsection{Comparison to SHAP}

We compare CADEX to the well known SHAP method \cite{Lundberg:2017} mentioned in section \ref{sec:explainability}. SHAP does not directly seek to find counterfactual explanations, but instead explain the effect of each input attribute on the resulting classification. Positive SHAP values are interpreted as increasing the likelihood of the observed classification, and vice versa for the negative values.

When CADEX produces a counterfactual explanation by modifying some attributes in the original input, we expect SHAP to have non-zero coefficients for the same attributes, since they are clearly important to the resulting classification. We have, however, observed that often that is not the case. We perform the comparison as follows. For each CADEX explanation found, we find the attributes which were modified, and count how many of them are zero in the SHAP coefficients of the original input. We used the SHAP implementation on github\footnote{https://github.com/slundberg/shap}, and used the kernel explainer with the training set as the background dataset. From the results in figure \ref{fig:shap} we see that in over\textbf{ 93\%} of the cases, at least one attribute modified by CADEX had a zero SHAP coefficient.

From the comparison we can learn that CADEX can find meaningful attributes to change in the input in order to get a counterfactual explanation, which are undetected and unexplained by SHAP.

\subsection{Transferability}
We have assumed so far that the bank in our scenario has used a neural network to assign the loan classification to its customers. We now consider the case where the bank has instead used another classifier, which is not a neural network. As shown by \cite{Szegedy:2014,Papernot:2017}, adversarial examples found using one model can transfer to another one trained on different but similar data, even if that model is not a neural network such as SVM, decision tree and logistic regression. We examined the transferability of CADEX explanations by training a random forest classifier on the same training set with 100 trees and the default scikit-learn parameters. Then, for each validation set item where the classifications of the neural net model and random forest model agreed, we checked how many explanations were indeed adversarial on the random forest model. We repeated the experiment 100 times with different random seeds. We found that on average, in \textbf{95.2\%} of the cases at least one CADEX explanation was adversarial on the random forest model, and in \textbf{87.6\%} of the time at least two. In total, \textbf{86.1\%} of all generated CADEX explanations were found to be adversarial on the random forest model.

This shows that the explanations are largely transferable. For future work, we can consider training more than one neural network model on the data, and to search all of them until a transferable explanation is found.

\section{Conclusion}
We have shown that CADEX is a robust method to produce counterfactual explanations. Such explanations are by nature highly understandable and actionable by people who receive them. We have demonstrated that CADEX is relatively easy to compute, and can be used to impose various domain and business constraints on the search process.

Going back to the bank scenario, we have shown how the hypothetical bank would benefit from having a way to generate such explanations to its customers. It can use the technique to allow a form of transparency where none exists today, without compromising itself. We believe that such approaches become crucial as machine learning models take a more active part in our daily lives, when we wish to be able to establish trust between the algorithm and the people it serves, and as the public demand for explainability increases.

\bibliographystyle{splncs04}
\bibliography{cadex}

\end{document}